\documentclass[conference]{IEEEtran}

\usepackage[utf8]{inputenc}
\usepackage[T1]{fontenc}

\usepackage{cite}
\usepackage{amsmath,amssymb,amsfonts}
\usepackage{algorithmic}
\usepackage{graphicx}
\usepackage{textcomp}
\usepackage{xcolor}
\usepackage{svg}

\usepackage[pagebackref,breaklinks,colorlinks]{hyperref}

\usepackage[capitalize]{cleveref}
\crefname{section}{Sec.}{Secs.}
\Crefname{section}{Section}{Sections}
\crefname{table}{Tab.}{Tabs.}
\Crefname{table}{Table}{Tables}
\crefname{figure}{Fig.}{Figs.}
\Crefname{figure}{Figure}{Figures}
\crefname{equation}{Eq.}{Eqs.}
\Crefname{equation}{Equation}{Equations}

\newcommand{\surl}[1]{\href{https://#1}{\nolinkurl{#1}}}
\newcommand{\uurl}[1]{\href{http://#1}{\nolinkurl{#1}}}
\newcommand{\enquote}[1]{"#1"}
\newcommand{\comment}[1]{}

\def\BibTeX{{\rm B\kern-.05em{\sc i\kern-.025em b}\kern-.08em T\kern-.1667em\lower.7ex\hbox{E}\kern-.125emX}}

\begin{document}

\title{
  A Study of Continual Learning Methods for Q-Learning
}

\author{
  \IEEEauthorblockN{
    1\textsuperscript{st} Benedikt Bagus
  }
  \IEEEauthorblockA{
    \textit{dept. of computer science} \\
    \textit{University of Applied Sciences Fulda} \\
    Leipzigerstr. 123, 36093 Fulda, Germany \\
    benedikt.bagus@cs.hs-fulda.de
  }
  \and
  \IEEEauthorblockN{
    2\textsuperscript{nd} Alexander Gepperth
  }
  \IEEEauthorblockA{
    \textit{dept. of computer science} \\
    \textit{University of Applied Sciences Fulda} \\
    Leipzigerstr. 123, 36093 Fulda, Germany \\
    alexander.gepperth@cs.hs-fulda.de
  }
}

\maketitle

\begin{abstract}
  We present an empirical study on the use of continual learning (CL) methods in a reinforcement learning (RL) scenario, which, to the best of our knowledge, has not been described before.
  CL is a very active recent research topic concerned with machine learning under non-stationary data distributions.
  Although this naturally applies to RL, the use of dedicated CL methods is still uncommon.
  This may be due to the fact that CL methods often assume a decomposition of CL problems into disjoint sub-tasks of stationary distribution, that the onset of these sub-tasks is known, and that sub-tasks are non-contradictory.
  In this study, we perform an empirical comparison of selected CL methods in a RL problem where a physically simulated robot must follow a racetrack by vision.
  In order to make CL methods applicable, we restrict the RL setting and introduce non-conflicting subtasks of known onset, which are however not disjoint and whose distribution, from the learner's point of view, is still non-stationary.
  Our results show that dedicated CL methods can significantly improve learning when compared to the baseline technique of \enquote{experience replay}.
\end{abstract}


\section{Introduction}\label{sec:int}
This article is in the context of continual reinforcement learning (CRL) \cite{Lesort2019, Khetarpal2020}, which describes the application of dedicated continual learning (CL) \cite{DeLange2019, Parisi2018} algorithms to reinforcement learning (RL) \cite{Sutton1998}.
Both are concerned with learning from non-stationary data distributions.
While assumptions in CL are quite varied, RL assumes a well-defined scenario.
RL is founded on Markov decision processes (MDPs), which are generally formalized as 5-tuple $M = \left<S, A, P, R, \rho_{0}\right>$.
$S$ and $A$ are the sets of all \textit{valid} states/actions, $P$ is the probability function of the transition $S \times A \to \mathcal{P}(S)$, with $P(\vec s_{t+1} = \vec s\,' | \vec s_{t} = \vec s, \vec a_{t} = \vec a)$ being the probability of transitioning into state $\vec s\,'$ if action $\vec a$ is taken in state $\vec s$.
$R$ is called reward function, which maps $S \times A \times S \to \mathbb{R}$ and provides the return signal $r_{t} = R(\vec s_{t}, \vec a_{t}, \vec s_{t+1})$.
Finally, $\rho_{0}$ is the distribution of the initial state.
Here, an \textit{agent} interacts with an \textit{environment}, trying to maximize a \textit{reward} $r_{t} \in \mathbb{R}$ signal by selecting the most appropriate \textit{action} $\vec a_{t} \in A$ based on an \textit{observation} $\vec o_{t}$, e.g., a camera signal, of the actual state $\vec s_{t} \in S$ following a policy $\pi$.
This feedback loop inherently defines an incremental manner of learning and implies appropriate models.
\begin{figure}[!ht]
  \centering
  \includegraphics[width=\linewidth]{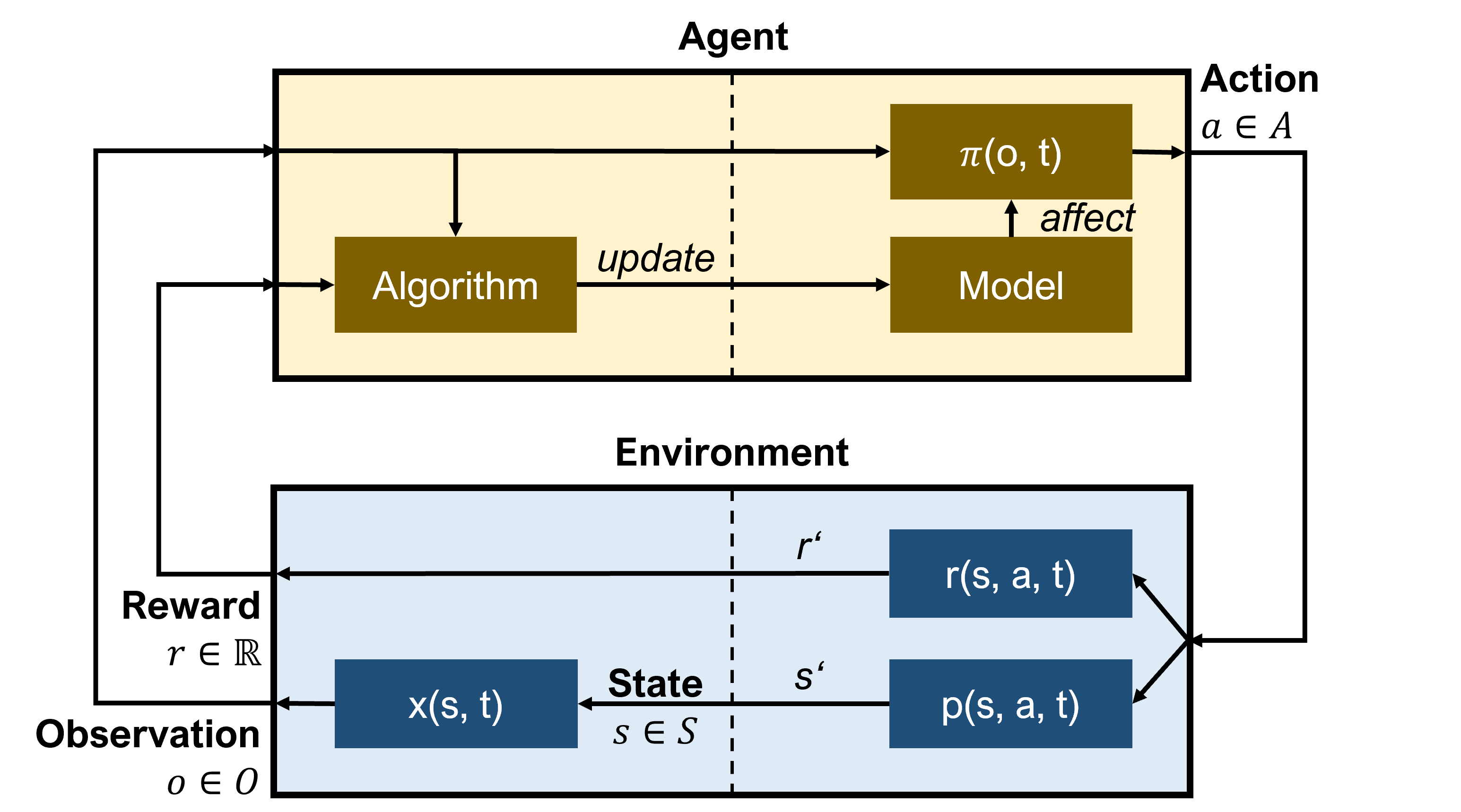}
  \caption{\label{fig:schema}
    Schematic representation of a RL control loop, which enables decision-making.
  }
\end{figure}

In this article we focus on \textit{Q-learning} \cite{Watkins1992}, an important flavor of RL, where the agent selects actions that maximize the expected future return $Q(s_{t}, a_{t})$, whose dependency on state and action must be acquired through learning.
\begin{equation}
  \begin{aligned}
    Q'(s_{t}, a_{t}) & \leftarrow (1 - \alpha) \cdot Q(s_{t}, a_{t}) + \\
    & + \alpha \cdot (r_{t} + \gamma \cdot \max_{a} Q(s_{t+1}, a))
  \end{aligned}
  \label{eq:qlearning_default}
\end{equation}
Q-Learning represents an \enquote{off-policy} approach, where exploration is commonly ensured by a $\epsilon$-greedy strategy and policy updates are immediately performed after an iteration.
The exploration ensures the discovery of diverse state-action combinations.
This is crucial for models like Q-tables, since they implement a defined assignment (lookup) between states and actions.

\begin{figure*}[!ht]
  \centering
  \includegraphics[width=\textwidth]{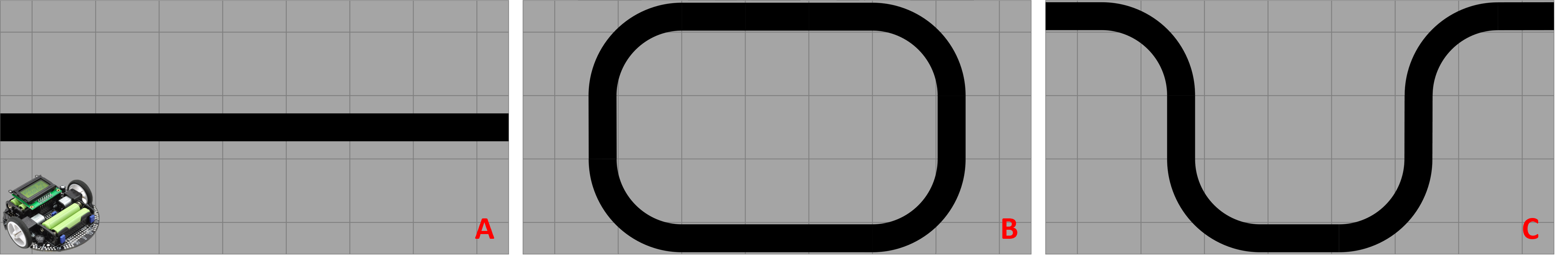}
  \caption{\label{fig:shifts}
    Illustration of environment shifts considered in this article.
    A simulated robot is trained to follow the black line in a succession of environments: A) straight line B) straight line + left curve C) straight line + left/right curve.
    Please note that all of these scenarios include concept drift as well from the point of view of the learner, due to the exploration of the state-action space.
  }
\end{figure*}

\subsection{Catastrophic forgetting in RL}
Conventional assumptions in machine learning are that data distributions are stationary and samples are iid.
In RL, these assumptions can be systematically violated in several ways:
\par\smallskip\noindent\textbf{Concept drift/shift}
Even if the environment is stationary, the observations will not be, at least not on the short timescale due to the ongoing exploration of the state-action space.
For example, when taking an action $\vec a\,'$ as a consequence of observation $\vec o\,'$, the obtained rewards may be in disagreement at first, since the environment is supposed to be stationary but not deterministic.
Similarly, the sampling of the state-action space is changing over time due to the decreasing influence of exploration, effectively creating another source of concept drift.
\par\smallskip\noindent\textbf{Environment shift}
In contrast to this, the environment itself may be non-stationary and subject to \textit{environment drift/shift} (see, e.g., \cite{Gepperth2016}), and thus automatically lead to non-stationary observations.
This includes the case where the agent encounters novel situations within the environment or is transferred to a completely different one.
Examples of environment shift are given in \cref{fig:shifts}.

Deep neural networks (DNNs) are typical machine learning models employed in deep reinforcement learning (DRL).
This raises the issue of catastrophic forgetting (CF) which especially DNNs are subject to (see, e.g., \cite{Pfuelb2019}) as a consequence of non-stationary data distributions.
Common workarounds include the use of \textit{experience replay} (ER), see \cite{Rolnick2018}.
In this approach, newly arriving samples are stored, e.g., by reservoir sampling, in a large buffer $\mathcal{M}$ with fixed size $M = |\mathcal{M}|$.
Instead of training the model directly, random mini-batches are drawn from the buffer $\mathcal{B} \sim \mathcal{M}$, which simulates a stationary data distribution.
This mitigates CF, but incurs a huge cost in memory.
In addition, reaction to concept drift/shift is delayed because new data samples will take a while before they are significantly represented in the buffer.
Simultaneously, if the sample selection is insufficient, the subset may not match the real distribution or underrepresented samples could be unbalanced even with buffer.

\subsection{Related work: CL approaches}\label{sec:int:cl}
The research field of continual learning (CL) investigates the problem of learning under non-stationary data distributions, see \cite{DeLange2019, Parisi2018} for reviews.
Systematic comparisons between different approaches to avoid CF are performed in, e.g., \cite{Kemker2017, Pfuelb2019}.
As discussed in \cite{Pfuelb2019, Prabhu2020, Zeno2018, Normandin2021}, many recently proposed methods demand specific experimental setups, which deviate significantly from applications.
In contrast to learning from stationary data, CL scenarios are very diverse \cite{Zeno2018, Normandin2021, Lesort2019, Khetarpal2020}, depending on what type of non-stationarity is assumed \cite{Lesort2021}.
Theoretical analyses of CL are presented in \cite{Chaudhry2018b, Knoblauch2020, Doan2021}, again underscoring the fact that a universally accepted definition of CL has not yet been reached.

A very common assumption in CL is the decomposition of a problem into several sub-tasks of \textit{stationary} statistics \cite{LopezPaz2017, Lesort2019}, which are locally iid.
Likewise, the onset of each sub-task is assumed to be known \cite{Zeno2018, Aljundi2018} which side-steps the issue of \textit{detecting} the boundaries.
Furthermore, sub-tasks are often assumed to be \textit{disjoint} \cite{Prabhu2020}: different sub-tasks contain different classes of the classification problem.
This implies that no re- or un-learning takes place, where samples from known classes would be assigned a different label in later sub-tasks \cite{Lesort2021}.

Among the proposed CL methods, three major directions may be distinguished according to \cite{DeLange2019}:
\par\smallskip\noindent\textbf{Parameter Isolation}
Parameter isolation methods aim at determining (or creating) a group of parameters that are mainly \enquote{responsible} for a certain sub-task.
CF is then avoided by protecting these parameters or adding new ones when training on successive sub-tasks, see \cite{Fernando2017, Mallya2017, Rusu2016}.
\par\smallskip\noindent\textbf{Regularization}
Regularization methods mostly propose modifying the loss function by including additional terms of criterions that protect knowledge acquired in previous sub-tasks, see \cite{Kirkpatrick2016, Aljundi2018, Lee2017}.
\par\smallskip\noindent\textbf{Replay}
Replay methods keep small subsets of real samples or train generators to reconstruct an arbitrary number of pseudo-samples afterwards.
CF can be circumvented by putting constraints on current sub-task training or by adding retained samples to the current sub-task, see \cite{LopezPaz2017, Chaudhry2018a, Chaudhry2019, Bagus2021, Shin2017, Gepperth2021}.

\subsection{Related work: Continual RL}\label{sec:int:crl}
As stated previously, CL is a broad topic and methods can be tailored to specific scenarios.
Transferring them to a more general domain such as RL is therefore a non-trivial objective.
First attempts to use CL methods in the domain of RL are performed in \cite{Lesort2019} and \cite{Khetarpal2020}.
The work of \cite{Lesort2019} gives detailed insights into important aspects and describes RL as a natural fit to CL.
In particular, the work of \cite{Khetarpal2020} elaborates extensively upon the application of CL in RL and describes essential criteria on an abstract level.
Both works introduce generic frameworks for CRL and discuss concepts, without implementing them.

In general, several frameworks for benchmarking RL have been presented, e.g., \cite{Brockman2016}.
However, only a few of them are formulated with CRL in mind \cite{Yu2019, Wolczyk2021, Powers2021}.
Due to the divergent formalism, established metrics as accuracy or forgetting/transfer measures \cite{LopezPaz2017, Chaudhry2018a} are inapplicable.
In RL, samples are not classified but assigned to Q-values.
However, new metrics for CRL are proposed in \cite{Lesort2019, Khetarpal2020} and \cite{Wolczyk2021, Powers2021}.

The works of \cite{Kessler2021, Zhang2021} combine CL and RL and suggest own approaches tailored to RL.
For example, \cite{Kessler2021} employ regularization-based models with multi-head outputs and \cite{Zhang2021} try to mitigate CF within single tasks by context detection.
However, established state-of-the-art CL methods have barely been reused and studied under generic conditions as provided by the RL domain.

\subsection{Mapping CL to RL}\label{sec:int:din}
RL is founded on an incremental formalism of a decision problem and offers an \enquote{real-world} application for CL.
However, when endeavoring to apply CL approaches in RL, there are restrictions that may be problematic due to deviating conditions:
\par\smallskip\noindent\textbf{Decomposition into sub-tasks}
One of the most wide-spread assumptions in CL concerns the decomposition of the learning problem into a sequence of \textit{sub-tasks}, with stationary data statistics and that they are locally iid.
In a stationary environment, the only entity that can be identified with a CL sub-task is an \textit{episode}, that is, a sequence between two terminal states.
It is unclear, though, whether statistics within an episode should be considered stationary.
Epochs themselves may contain context changes and would therefore violate the common definition of CL sub-tasks.
\par\smallskip\noindent\textbf{Overlapping and contradictory sub-tasks}
Almost universally, sub-tasks in CL are supposed to be disjoint.
As a consequence, many approaches try to identify parameters that are important for certain sub-tasks, or to directly dedicate parts of the ML model to certain sub-tasks.
Typical representatives are regularization approaches like \cite{Kirkpatrick2016, Aljundi2018, Lee2017}, but also parameter isolation methods like \cite{Fernando2017, Mallya2017, Rusu2016}.
In RL, episodes do not contain consistent labels or separated classes because they are treated differently.
Each \enquote{label} consist of non-one-hot always changing Q-Values, whereby each sample (state) can be assigned to one discrete action.
This will lead to difficulties if there exist contradicting, common samples in different sub-tasks, a frequent situation in RL, especially when sub-tasks are defined by RL episodes.
Even environment shifts would not alleviate this problem. 
Therefore, virtually all CL methods will encounter problems in such cases.
\par\smallskip\noindent\textbf{Small number of sub-tasks}
Most CL models tacitly assume that the number of sub-tasks is small, as they need to store significant amounts of data for each sub-task.
As an example, EWC \cite{Kirkpatrick2016} needs to store a Fisher Information Matrix (FIM) plus all trainable parameters of a DNN for each sub-task, apart from the fact that the loss function picks up a new term for each additional sub-task.
Parameter isolation methods face similar problems: either they need to track which parameter is important for which sub-task, or new structures must be added to the ML model for each sub-task, both of which are costly in terms of memory and computational effort.
Exceptions are GEM and A-GEM \cite{LopezPaz2017, Chaudhry2018a}, since they just store a small percentage of samples per sub-task, and thus the number of sub-tasks can be large.
Pure rehearsal-based methods such as NSR\texttt{+} \cite{Bagus2021} pursue a similar strategy, with similar advantages and high computational efficiency (storing a few samples is cheap).
Pseudo-rehearsal methods such as, e.g., \cite{Shin2017}, do not store samples but train a generator.
This can be done for any number of sub-tasks, but incurs a huge \textit{fixed} computational and memory cost, and is therefore slightly less suited.
\par\smallskip\noindent\textbf{Sharp and known boundaries}
In addition, CL sub-task-boundaries are usually assumed to be sharp, and their onset known.
If common CL approaches are to be applied, transitions between sub-tasks must be detected.
However, sub-task onsets in RL may be smooth to a point where even the concept of a sub-task becomes questionable.

\subsection{Goals and contribution}\label{sec:int:goa}
In order to apply CL methods to RL, we created an RL scenario where non-stationarities are mainly characterized by environment shifts, see \cref{fig:shifts}. 
Sub-tasks are overlapping but non-contradictory (e.g., high rewards in situations that previously obtained low rewards are non-existing) and their onsets are assumed to be known.
Although we use only a small number of tasks, we chose CL methods that could scale well even for a moderate amount of sub-tasks: Gradient Episodic Memory (GEM) \cite{LopezPaz2017} and Averaged Gradient Episodic Memory (A-GEM) \cite{Chaudhry2018a} and Naive Sample Rehearsal Plus (NSR\texttt{+}) \cite{Bagus2021}.
As a baseline for RL performance, a matrix-based model and the conventional ER approach are investigated.

We present several novel contributions:
\begin{itemize}
  \item Review of existing CL methods w.r.t. suitability for reinforcement learning
  \item Adaptation of CL reference methods to the domain of RL
  \item Experimental comparison of various CL methods in a realistic RL scenario
\end{itemize}

\section{Methods}\label{sec:met}
\par\smallskip\noindent\textbf{Simulation environment}
We use Robot Operating System 2 (ROS2) and Gazebo 11\footnote{\surl{docs.ros.org} and \surl{gazebosim.org} for further information} to physically simulate and control a line-following robot.
ROS2 is a common middleware framework for robotics, which allows decoupled communication between multiple heterogeneous entities (nodes).
We employ it to communicate with the simulated robot by sending control signals and receiving sensor data.
Gazebo 11 is a physics-based simulator for ROS2 and includes plugins for various pre-defined types of sensors as well as actuators.
Our simulations consist of at least one simulated world (environment), each of which contains a particular racetrack, and the robot model (interacting agent), which can observe via its sensors and perform actions via its actuators.
Control signals and sensor readings are transmitted at a defined frequency (5 Hz).

\par\smallskip\noindent\textbf{Racetracks}
All racetracks consist of separate lines on a plane, which can have an arbitrary course, but never crosses.
Subsections of them can therefore be divided into three categories: straight, left turn or right turn.
Context changes can be reasonably interpreted as transitions between these distinct categories, see \cref{fig:shifts}.

\begin{figure}[!ht]
  \centering
  \includegraphics[width=0.75\linewidth]{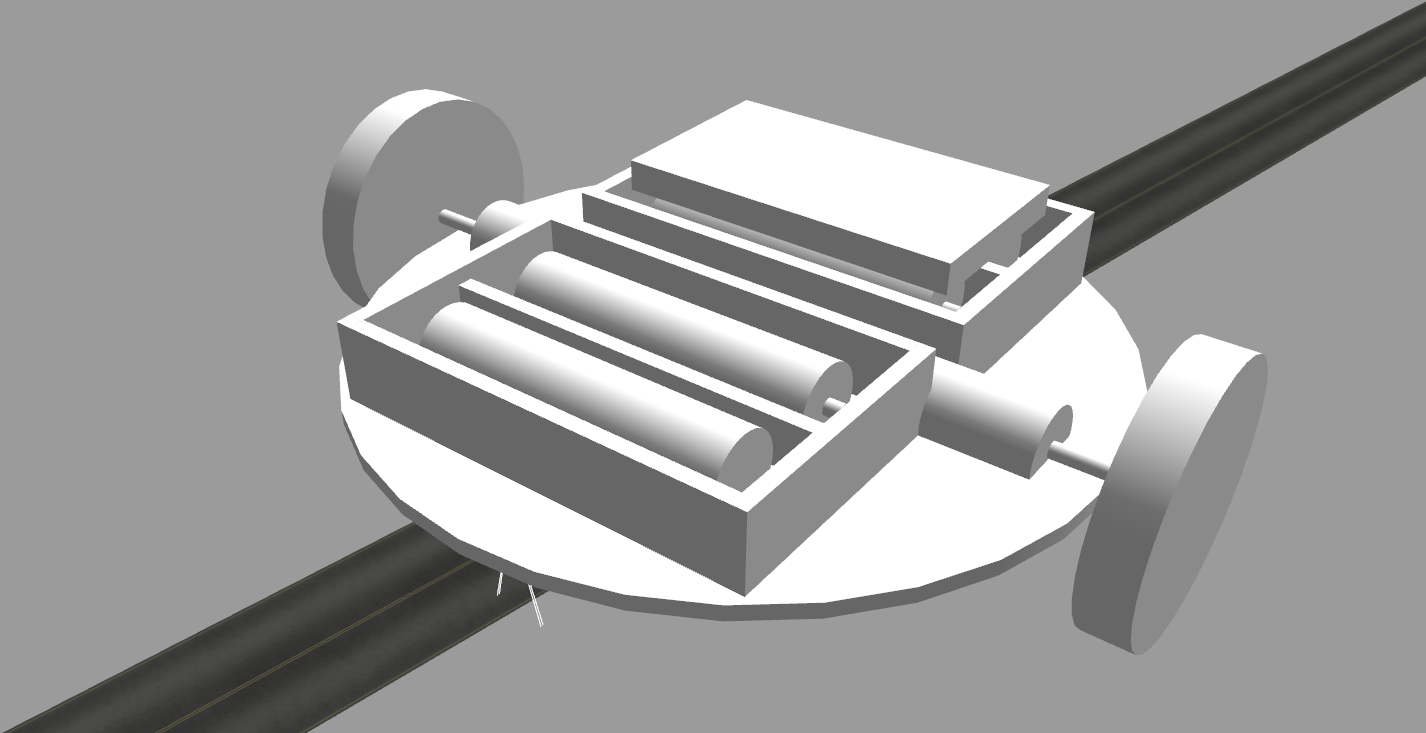}
  \caption{\label{fig:robot}
    The simulated robot, which closely models the popular $3\pi$ robot.
  }
\end{figure}
\par\smallskip\noindent\textbf{Simulated robot}
The simulated robot is modeled after the popular $3\pi$ robot from Pololu robotics, see \cref{fig:robot}.
It is controlled by a differential drive, with two wheels (radius: $\approx 1.55\,cm$, separation: $\approx 9\,cm$) driven by independent motors.
In addition, there is a passive caster wheel for balancing.

\begin{table}[!ht]
  \centering
  \caption{\label{tab:robot}
    Key data of the used robot, which is a replica of the Pololu $3\pi$.
  }
  \begin{tabular}{c|c|c|c}
    \hline
    \textbf{Height} & \textbf{Width} & \textbf{Length} & \textbf{Weight} \\
    \hline
    $\approx 3\,cm$ & $\approx 9.5\,cm$ & $\approx 9.5\,cm$ & $\approx 135\,g$ \\
    \hline
  \end{tabular}
\end{table}

\par\smallskip\noindent\textbf{Action space}
The action space consists of 9 discrete actions $a(t)$: the three basic actions (drive straight, turn left or turn right) executed at three different speeds.
An action is a 2-tuple containing a value for the speed of each wheel (left and right).
The individual wheel speeds for all actions are given in \cref{tab:actions}.
\begin{table}[!hb]
  \centering
  \caption{\label{tab:actions}
    Discrete action space, wheel speeds are given in meter per second ($\frac{m}{s}$).
  }
  \resizebox{\linewidth}{!}{
    \begin{tabular}{c|ccc|ccc|ccc}
      \hline
      \textbf{Index} & \textbf{1} & \textbf{2} & \textbf{3} & \textbf{4} & \textbf{5} & \textbf{6} & \textbf{7} & \textbf{8} & \textbf{9} \\
      \hline
      \textbf{Left speed} & $0.00$ & $0.05$ & $0.10$ & $0.00$ & $0.00$ & $0.05$ & $0.05$ & $0.10$ & $0.10$ \\
      \textbf{Right speed} & $0.00$ & $0.05$ & $0.10$ & $0.05$ & $0.10$ & $0.10$ & $0.00$ & $0.00$ & $0.05$   \\
      \hline
      \textbf{Type} & \multicolumn{3}{|c|}{\textbf{Straight}} & \multicolumn{3}{|c}{\textbf{Left}} & \multicolumn{3}{|c}{\textbf{Right}} \\
      \hline
    \end{tabular}
  }
\end{table}

\par\smallskip\noindent\textbf{Reward}
The reward signal is calculated based on the deviation $d$ (in pixels) of the left edge of the line to the center of the image, which is assumed to have width $W$:
\begin{equation}
  \begin{aligned}
    r(t) & = 0.5 - \left|\frac{d - \frac{W}{2}}{\frac{W}{2}}\right| \text{, with } d \in [0, W] \\
  \end{aligned}
  \label{eq:reward}
\end{equation}
The deviation is computed by image processing, assuming that the line is significant darker than the remaining ground plane.
This measure provides a \enquote{dense} reward, which is computable immediately for each iteration in our feedback loop.
The default range of the signal is within $[-0.5, +0.5]$, but terminal states and neutral actions (no movement) are penalized by a value of $-1.0$.
A terminal state always occurs when image processing is unable to detect the left edge of the line in the image.
The reward function does not reward or penalize different speeds, except for the neutral action (no movement).

\par\smallskip\noindent\textbf{State Space}
The agent observes its environment by a downward directed camera ($5 \times 100 \times 3$ pixels).
Matrix controllers are preprocessing such data to produce a scalar state $s(t)$, which is realized by the deviation $d$.
All other controllers stack the last $n=5$ received images to obtain a state representation $\vec s(t)$ of dimension $5n \times 100 \times 1$.

\par\smallskip\noindent\textbf{Episode definition}
The investigated scenario describes a control problem with a theoretically infinite length of episode.
A single episode can therefore be of arbitrarily length and trajectories consist of varying numbers of samples.
An episode ends only if a terminal (\textit{invalid}) state is reached, and the agent must be reset in order to start over again.

\par\smallskip\noindent\textbf{Sub-task definition}
Each new sub-task consists of segments in which novel skills must be acquired, but also consist of sections from previous sub-tasks, which require the adoption of already learned ones.
In this work, for the time being, only a deliberate change of the racetrack (referred as environment shift) will be considered as a sub-task.
This relaxation permits the application of existing state-of-the-art methods.

\subsection{CL approaches}\label{sec:met:inv}
We investigate three different CL approaches.
All of them are replay-based methods and should replace the original ER.

\par\smallskip\noindent\textbf{GEM and A-GEM}
Gradient Episodic Memory (GEM) \cite{LopezPaz2017} and Averaged Gradient Episodic Memory (A-GEM) \cite{Chaudhry2018a} are both constraint-based replay approaches.
They project the current loss gradient onto loss gradients computed from samples in the buffer, if deviations are judged to be excessive.
GEM and A-GEM share a common principle, but differ in the way of computing loss gradients from buffered samples.
The primary parameter of (A-)GEM is the size $M$ of buffer $\mathcal{M}$.
GEM and A-GEM do not define a specific sample selection strategy and propose to store randomly $m = \frac{M}{t}$ samples of the current sub-task.
However, the selection requires information about sub-task boundaries in order to assign distinct partitions per sub-task.

\par\smallskip\noindent\textbf{NSR\texttt{+}}
Naive Sample Rehearsal Plus (NSR\texttt{+}) \cite{Bagus2021} is another replay approach, which can be executed with the information about task boundaries, but unlike the (A-)GEM the sample-selection does not depend on it.
NSR\texttt{+} has two parameters: the size $M$ of buffer $\mathcal{M}$ as well as the ratio $r$ of replayed samples.
The selection strategy keeps the $M$ worst-performing samples in a buffer, for this the buffer is evaluated as a separate mini-batch and compared with the losses of the current mini-batch.

\subsection{Models}\label{sec:met:mod}
Q-learning is performed using matrix-type and DQN-type models.
The matrix-type model is trained either by the original update rule \cite{Watkins1992} or by the update rule of speedy Q-learning \cite{Azar2011}.
DQN-type models are trained either by the standard DQN update rule \cite{Mnih2013} or by double Q-learning \cite{Hasselt2015}.

\par\smallskip\noindent\textbf{Q-tables}
First of all, matrix-type models (Q-tables) are used to obtain reinforcement learning in the original manner and to compare with the other models.
A Q-table consists of $(|S| \times |A|)$ entries and is able to map discrete states onto discrete actions.
Each entry (state-action tuple) represents the expected return as Q-value $Q(s(t), a(t))$ if this action would be chosen in the given state.
Therefore, high values are to be preferred, and the maximum value is to be determined via the respective state.
Like all Q-learning models, Q-tables are updated after each single time step.
However, only the value of the last state-action tuple will be updated, thus all other values remain unaffected.

\par\smallskip\noindent\textbf{DQN}
Instead of matrices, DNNs can be used as back-end to obtain deep reinforcement learning (DRL) by deep Q-networks (DQNs).
This has multiple advantages: the state space can be continuous, the state can be in context, the state can be consisted of multi-modal data and the model can generalize between samples.
The type of network is freely selectable and independent of the chosen algorithms.
Some networks are more or less suitable for different scenarios, for simplicity we use conventional convolutional neural networks (CNNs).
Like Q-tables, DQNs perform an update after every time step but contrary this involves modifications to all trainable parameters ($\theta$).
Each update is w.r.t. the current sample or mini-batch, and can therefore amplify the effect of catastrophic forgetting (CF) if gradients are opposite.

\subsection{RL Sub-tasks}\label{sec:met:tas}
\begin{table}[!ht]
  \centering
  \caption{\label{tab:subtasks}
    Sub-task definitions for continual RL as employed in this article.
    Each sub-task consists of a total of 50,000 time steps and adds new capabilities.
  }
  \begin{tabular}{c|ccc}
    \hline
    \textbf{ID} & \textbf{1} & \textbf{2} & \textbf{3} \\
    \hline
    \textbf{Type} & straight & zero & slalom \\
    \textbf{Start it.} & $0$ & $50,000$ & $100,000$ \\
    \hline
  \end{tabular}
\end{table}

CL problems are typically divided into individual \textit{sub-tasks}, following the terminology from \cite{Pfuelb2019}.
Frequently, class labels are used as a basis to group samples into disjoint sub-tasks, since most work on CL is in the context of supervised learning.
Since class labels do not exist in RL, this definition of sub-tasks cannot be used for RL.
Instead, we focus on environment shifts for defining RL sub-tasks, which we create by manipulating the track definitions (\cref{fig:shifts}) at defined intervals, see \cref{tab:subtasks}.
In contrast to the main body of CL, sub-tasks as understood here are not disjoint.
More precisely, each new sub-task requires additional skills to obtain high rewards, but also re-uses skills acquired in previous sub-tasks.
For example, an agent solving the second sub-task can re-use the skill to drive on a straight line (acquired in the first sub-task), but needs to acquire the skill to execute curves to the left in addition to that.
\\
By implementing sub-tasks as synthetic environmental shifts, see \cref{fig:shifts}, we control when they occur and make this information available to the RL process.
Concretely, we reset the parameter $\epsilon$ which controls exploration behavior to a value of $\frac{1}{t}$ for each sub-task, after which the normal decaying-$\epsilon$-greedy is performed as usual.
Moreover, the utilized Cl methods can use this information to, e.g., generate new partitions.
In a fully realistic RL application, the agent would have to discover such sub-task boundaries by itself.

\subsection{Benchmarking}\label{sec:met:ben}
While in supervised learning the terms benchmarks and datasets are synonymous, in reinforcement learning a distinction must be made.
Henceforth, the term \enquote{benchmark} describes a possibility to evaluate the current policy ($\pi$) of an agent against certain metrics to capture its performance,  but without predetermined samples to serve as a \enquote{dataset}.
The fundamental issue here is that the reward is not only a function of the current observation, but also of a previous action.
In the same way, the current observation depends on previous actions.
In order to perform an offline evaluation of a learned policy, this policy would be required to take a predetermined sequence of actions, which in turn would prevent a thorough evaluation of the policy whose goal, after all, is to \textit{select} appropriate actions.

Any evaluation therefore has to be performed \textit{within} an environment using a fixed policy (i.e., by setting the learning rate to $0$ and suppressing exploration), in contrast to the offline evaluation of classifiers that is common in CL.
We choose to perform benchmarking directly after training on a sub-task is completed.

\subsection{Metrics}\label{sec:met:met}
We are using the history of measured rewards to evaluate learned policies.
An elementary quantity of our evaluation is the sum over all rewards $\Sigma_{e}$ received during a single RL episode $e$.
We chose summation over averaging since rewards are bounded, and the sum thus reflects not only the total obtained reward, but also the length of the episode.
By default, we plot the $\vec \Sigma_{t} = \Sigma_{e},\,\forall e \in t$ over the whole sub-task $t$ to obtain a visualization with intuitive meaning.
We can condense the information contained in $\vec \Sigma_{t}$ further by plotting an exponentially smoothed version, or even averaging them to obtain a scalar quality measure $\Sigma_{t}$ (which is strictly of a \enquote{the higher the better} nature).
\\
As already described in \cref{sec:met:ben}, each action during evaluation is chosen w.r.t. the current \textit{static} policy, without choosing some of them randomly, e.g., by a $\epsilon$-greedy strategy.

\section{Experiments}\label{sec:exp}
Different agents have to perform our benchmark in real-time.
Updates are executed in a strictly online fashion during the entire simulation.
After each sub-task, the respective policy of the agent is stored to compare experiments successively.
Finally, these agents should at least beat the baselines.

\subsection{Baselines}\label{sec:exp:bas}
In RL, baselines are closely linked to the respective benchmark and are non-transferable.
Furthermore, the variance between isolated runs can vary significantly, especially if the number of iterations is limited.
For comparison, all methods should be applied under identical conditions.
We chose state-of-the-art methods like Q-tables and DQNs with ER as possible lower bounds.
Both are commonly used in the domain of (D)RL, whereas we apply them here in the context of continual RL (CRL).
\begin{table}[!hb]
  \centering
  \caption{\label{tab:baselines}
    Performance of the best hyper-parameter setup for all baselines, represented by $\Sigma_{t}$ for each sub-task.
  }
  \resizebox{\linewidth}{!}{
    \begin{tabular}{cc|cccc}
      \hline
       & & \textbf{Sub-task 1} & \textbf{Sub-task 2} & \textbf{Sub-task 3} & \textbf{Overall} \\
      \textbf{Baseline} & \textbf{Type} & $\Sigma_{1}$ & $\Sigma_{2}$ & $\Sigma_{3}$ &  \\
      \hline
      \textbf{Q-tables} & history & $-911.833$ & $-5973.567$ & $-383.633$ & $-7269.033$ \\
       & last policy & $-164.333$ & $-622.000$ & $17.600$ & $-768.733$ \\
      \textbf{DQNs + ER} & history & $-1744.880$ & $-2777.807$ & $-996.160$ & $-5518.847$ \\
       & last policy & $-57.053$ & $-256.953$ & $-55.227$ & $-369.233$ \\
      \hline
    \end{tabular}
  }
\end{table}
Both baselines (\cref{tab:baselines}) achieve predominantly negative scores, while the second sub-task is performing the worst.
Negative scores indicate, that an agent is not able to keep the left edge of the line centered.
This does not have to result in a loss of the line from its field of view, but rather in an inconvenient driving style.
The benchmark itself seems already challenging, as the negative first sub-task indicates, regardless of environmental shifts being performed.
To rule out the possibility of having too few iterations to acquire needed skills, we additionally run these experiments $10$ times longer for $500,000$ iterations per sub-task.
\\
The runs differ because they are influenced by numerous \textit{random} factors, e.g., content-related aspects as exploration or initial decisions and also technical aspects as transmission time or calculation time.
However, the presented values are averaged over multiple runs (min $3$) to be as representative as possible.

\subsection{Results}\label{sec:exp:res}
The results of the investigated CL methods are shown in \cref{tab:outcomes}.
Compared with the baselines of \cref{tab:baselines}, each individual method results in a significant higher score.
But here, too, the second sub-task performs the worst.
\begin{table}[!ht]
  \centering
  \caption{\label{tab:outcomes}
    Performance of the best hyper-parameter setup for all CL method, represented by $\Sigma_{t}$ for each sub-task.
  }
  \resizebox{\linewidth}{!}{
    \begin{tabular}{cc|cccc}
      \hline
      & & \textbf{Sub-task 1} & \textbf{Sub-task 2} & \textbf{Sub-task 3} & \textbf{Overall} \\
      \textbf{Baseline} & \textbf{Type} & $\Sigma_{1}$ & $\Sigma_{2}$ & $\Sigma_{3}$ &  \\
      \hline
      \textbf{GEM} & history & $3143.780$ & $-1899.703$ & $2303.273$ & $3547.350$ \\
      & last policy & $450.300$ & $-109.193$ & $273.873$ & $614.980$ \\
      \textbf{A-GEM} & history & $5568.810$ & $-4470.237$ & $2663.120$ & $3761.693$ \\
      & last policy & $431.017$ & $-386.953$ & $316.243$ & $360.307$ \\
      \textbf{NSR\texttt{+}} & history & $4513.033$ & $-4049.577$ & $2746.470$ & $3209.927$ \\
      & last policy & $664.837$ & $-352.727$ & $304.730$ & $616.840$ \\
      \hline
    \end{tabular}
  }
\end{table}
Our results demonstrate that RL methods as Q-learning converges faster in combination with CL methods, since the first sub-task already has a high positive value.
New skills after sub-task changes (realized by environment shifts) are also adapted faster than with conventional ER approaches.
While ER trains over mini-batches, Q-Tables and the CL methods use single samples per update.

\begin{figure}[!ht]
  \centering
  \includesvg[width=\linewidth]{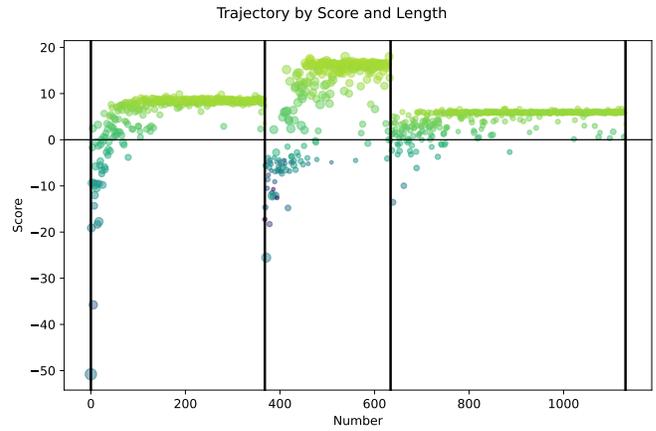}
  \caption{\label{fig:score}
    All $\Sigma_{e}$s of an outstanding A-GEM run.
    Data points are scaled by episode length and colored by the average.
    Sub-task boundaries are marked as vertical lines.
  }
\end{figure}
If an experiment has too few iterations (time steps), the results depends strongly on the learning process initiation, which leads to divergent or almost inconsistent results.
In addition to already mentioned factors, early good decisions in particular favor this, since wrong decisions merely rule them out but do not offer a selectable option.
The run visualized in \cref{fig:score} is nevertheless a representative trajectory of A-GEM experiments, as they are generally able to adapt new sub-tasks quickly.
Only the second sub-task performed in this case significantly better than average.

\section{Discussion}\label{sec:dis:kfi}
The experiments presented in \cref{sec:exp:res} allow to draw the following conclusions:
\par\smallskip\noindent\textbf{CL methods can be transferred to the RL domain}
Although we consider a simplified RL scenario here, where concept drift occurs only at sub-task boundaries, we nevertheless showed that the assumptions made by several dedicated CL methods can be made compatible with RL.
A significant percentage of CL methods were, however, excluded from our considerations due to memory or scalability issues, see \cref{sec:con} for a justification.
\par\smallskip\noindent\textbf{CL methods can improve RL performance}
W.r.t. to the baseline of experience replay, we observe that CL methods achieve faster convergence on single sub-tasks.
This shows that more knowledge is retained by CL methods.
Experience replay converges, but struggles with environment shifts, since it takes some time until the replay buffer is sufficiently populated with new sub-task samples.
\par\smallskip\noindent\textbf{Sample selection is essentially}
Both investigated CL methods perform replay, i.e., they re-use stored samples from past sub-tasks.
Our investigation indicates that sample selection is an essential ingredient, especially in RL.
To illustrate this, we recall that sub-tasks in RL (whatever their definition) are never disjoint.
Even very rudimentary sample selection strategies, like selecting the worst-performing samples as we did for NSR\texttt{+}, increases the chances of storing only samples characteristic to a certain sub-task.

\section{Conclusion and Outlook}\label{sec:con}
In this article, we consider a RL problem with known, hard sub-task boundaries.
Sub-tasks are not disjoint but overlapping, and no contradictions occur between sub-tasks, meaning that for the same sensory state, the Q-values do no change between sub-tasks.
In this somewhat restricted RL setting, we could show that two common CL methods, (A-)GEM and NSR\texttt{+} can outperform the common RL baseline of experience replay.
In this setting, we expect other CL methods to be applicable as well, e.g., generative replay and GMR.
However, several open issues remain and will need to be addressed in future work on CRL.
\par\smallskip\noindent\textbf{Known sub-task boundaries}
First of all, virtually all existing CL methods depend on the knowledge of sub-task boundaries, but are themselves incapable of detecting them.
Thus, CL algorithms with outlier detection capacity should be beneficial in this context, or else an additional mechanism performing the detection of sub-tasks.
However, the case of hard sub-task boundaries, as we introduced it here by the concept of environment shifts, is not a common one in RL.
Rather, data statistics change gradually, at least from the learner's point of view.
The absence of hard sub-tasks may exclude many CL approaches (e.g., constraint-based methods), or else force a major redesign.
\par\smallskip\noindent\textbf{Contradictions}
Any RL agent is likely to encounter conflicting data statistics, where the same state-action tuples will get assigned very different rewards as a consequence of, e.g., concept drift.
Most CL approaches implicitly assume that knowledge from previous sub-tasks should be preserved, and not modified.
This is no longer possible if systematic contradictions occur.
Identifying or designing a CL method that can work with conflicting data statistics will be a challenge to solve.
\par\smallskip\noindent\textbf{Scalability}
The scaling behavior of many CL approaches w.r.t. time and memory is unfavorable.
Some approaches require the storage of a complete model for each new sub-task, whereas replay methods must generate an ever-increasing amount of previous data for each new sub-task.
Since RL is often conducted over long time scales, with a succession of many sub-tasks, employing a scalable CL approach will be vital.

\section*{Supplementary Material}\label{sec:sup}
\subsection{DQN back-end}
\Cref{tab:cnn_network} presents detail about the back-end, we employed for all experiments.
The model is an intentionally small CNN and consists of two convolutional layers as well as two fully connected layers and the output layer.
For activation, ReLU is applied and after each convolution a maxpooling is performed.
Compared to the state-action space of our Q-tables ($33 \times 9 = 297$), this model ($48,613$) seems already overparameterized.

\begin{table}[!ht]
  \centering
  \caption{\label{tab:cnn_network}
    Summary of the CNN back-end architecture.
    The number of trainable parameters is $48,613$ in total.
  }
  \begin{tabular}{l|l|l|l}
    \hline
    \textbf{layer} & \textbf{type} & \textbf{properties} & \textbf{parameters} \\
    \hline
    input & identity & $5 \times 100 \times 3$ & \\
    conv 1 & convolutional & filters: 4 & $368$ \\
     & & kernel size: (3, 5) &  \\
     & & strides: (1, 1) &  \\
    maxp 1 & pooling & pool size: (1, 2) &  \\
    relu 1 & activation & ReLU &  \\
    conv 2 & convolutional & filters: 8 & $1,936$ \\
     & & kernel size: (3, 5) &  \\
     & & strides: (1, 1) &  \\
    maxp 2 & pooling & pool size: (1, 2) &  \\
    relu 2 & activation & ReLU &  \\
    \textit{flatten} & \textit{reshape} & \textit{352} &  \\
    fc 1 & dense & 100 & $35,300$ \\
    relu 3 & activation & ReLU &  \\
    fc 2 & dense & 100 & $10,100$ \\
    relu 4 & activation & ReLU &  \\
    output & dense & 9 & $909$ \\
    linear & activation & linear &  \\
    \hline
  \end{tabular}
\end{table}

\subsection{Hyper-parameters}
The list of all hyper-parameters we used in our specific RL setup is given by \Cref{tab:hyper-parameters}.
For better clarity, these have been grouped by subject.
The majority is needed to perform Q-learning or RL in general, only a few of them belong to CL methods.
We chose the sizes of CL buffers to be $\frac{1}{10}$-th of the ER buffer sizes.

\begin{table}[!hb]
  \centering
  \caption{\label{tab:hyper-parameters}
    List of all hyper-parameters and their corresponding values, we consider in this article.
  }
  \begin{tabular}{l|l|l}
    \hline
    \textbf{subject} & \textbf{name} & \textbf{values} \\
    \hline
    generic & transmission frequency & 5 Hz \\
    generic & state space complexity & 33 tuples (Q-tables) \\
     & & 500 pixels (DQNs) \\
    generic & action space complexity & 9 tuples \\
    generic & sequence size & 1 sample (Q-tables) \\
     & & 3 samples (DQNs) \\
    \hline
    task & iterations sub-tasks (train) & 50,000 \\
    task & iterations sub-tasks (eval) & 5,000 \\
    \hline
    algorithm & type & original, speedy (Q-tables) \\
     & & original, double (DQNs) \\
    algorithm & learning rate & 0.5, 0.1 (Q-tables) \\
     & & 1e-2, 1e-3 (DQNs) \\
    algorithm & discount factor & 0.75 \\
    algorithm & repeat action & 1 \\
    \hline
    model & update frequency & 1 \\
    model & mini-batch size & 1 (non ER) \\
     & & 8, 16 (ER) \\
    model & repeat update & 1 (non CL) \\
     &  & 3 (CL) \\
    model & loss function & Huber \\
    model & optimizer & SGD \\
    \hline
    $\epsilon$-greedy & strategy & non-linear decay \\
    $\epsilon$-greedy & start & $\frac{1}{t}$ \\
    $\epsilon$-greedy & stop & 0.005 \\
    $\epsilon$-greedy & step & 0.001 \\
    \hline
    \hline
    ER buffer & sample selection & reservoir sampling \\
    ER buffer & buffer size & 10,000, 30,000 \\
    \hline
    (A-)GEM & averaged & yes, no \\
    (A-)GEM & buffer size & 1,000, 3,000 \\
    (A-)GEM & memory strength & 0.5 \\
    \hline
    NSR\texttt{+} & buffer size & 1,000, 3,000 \\
    NSR\texttt{+} & replay ratio & 1.0 \\
    NSR\texttt{+} & sample selection & worst performing \\
    \hline
  \end{tabular}
\end{table}

\subsection{Policy evaluation}
\begin{figure}[!ht]
  \centering
  \includegraphics[width=\linewidth]{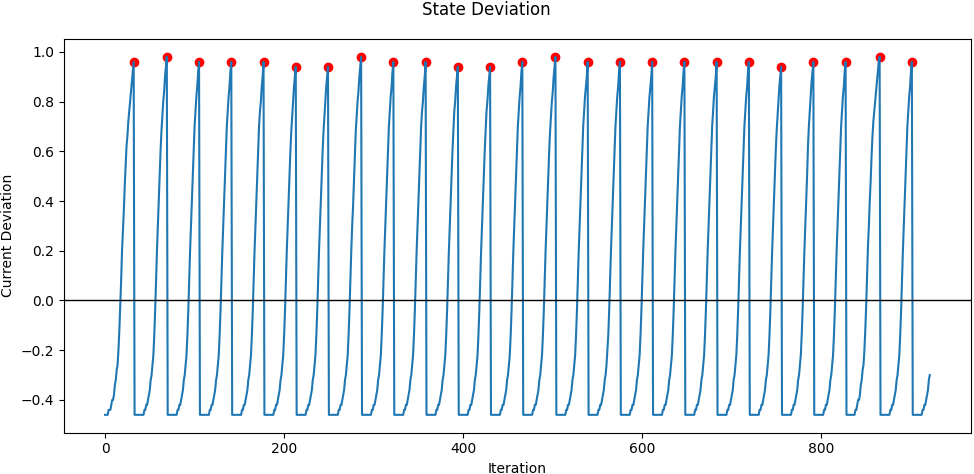}
  \caption{\label{fig:states_1}
    The agent's deviation from the line, evaluated before training.
  }
\end{figure}

\begin{figure}[!ht]
  \centering
  \includegraphics[width=\linewidth]{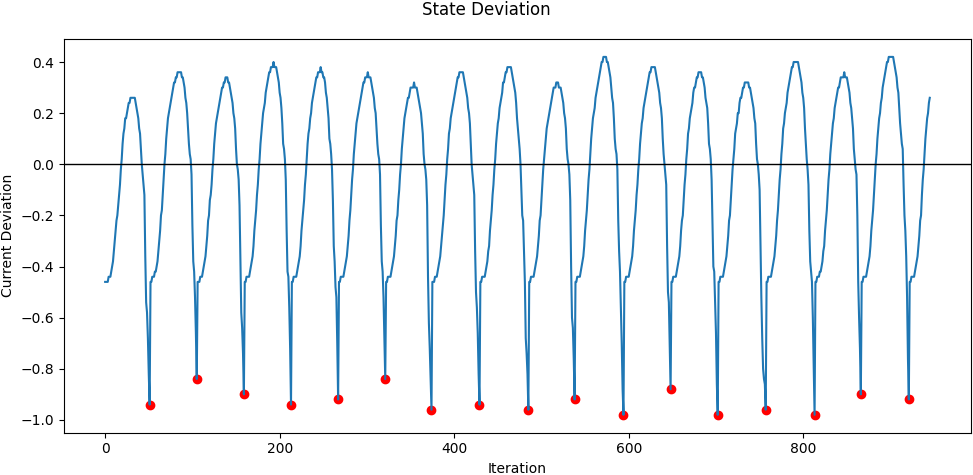}
  \caption{\label{fig:states_2}
    The agent's deviation from the line, evaluated early after training.
  }
\end{figure}

\begin{figure}[!ht]
  \centering
  \includegraphics[width=\linewidth]{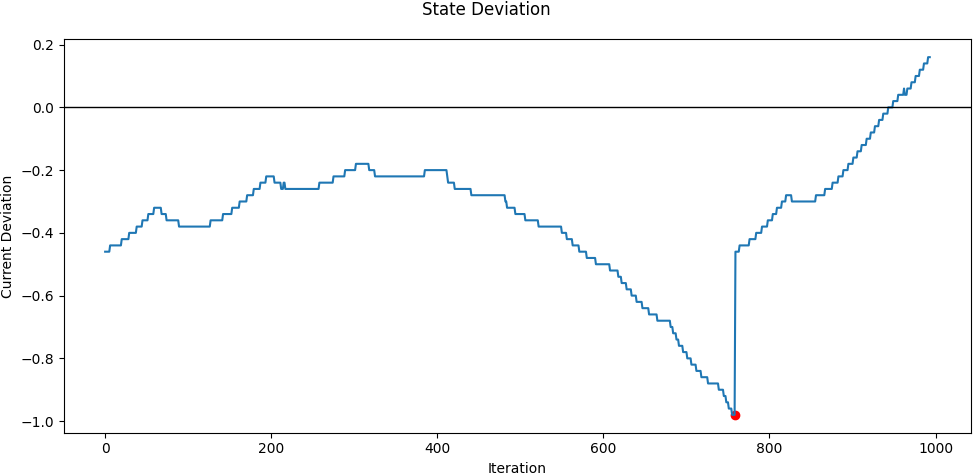}
  \caption{\label{fig:states_3}
    The agent's deviation from the line, evaluated later after training.
  }
\end{figure}

\Cref{fig:states_1,fig:states_2,fig:states_3} visualize the evaluation of line deviation within the observed images for $1,000$ iterations.
If the agent drives perfectly centered over the left edge of the line, the deviation will be $0$.
If the agent deviates to the left, the edge is moving to the right in the received image accordingly and vice versa.
Finally, the boundary of $-1$ encodes the edge in the leftmost position within the image, and $+1$ thus the rightmost position.
\\
However, \cref{fig:states_1} shows an evaluation of an untrained agent, which always drives to the left and loses the track immediately.
The evaluation of the same agent after some train steps is given in \cref{fig:states_2}.
Now, if the agent deviates too much to the left, he has learned to countersteer to the right, but still loses the track quickly.
Nevertheless, this represents the first acquired skill of the investigated agent.
Finally, \cref{fig:states_3} visualizes this agent at an even later stage.
It can be seen that actions are now much more finely motorized than before, but sometimes the agent still loses the track.

\begin{figure}[!ht]
  \centering
  \includesvg[width=\linewidth]{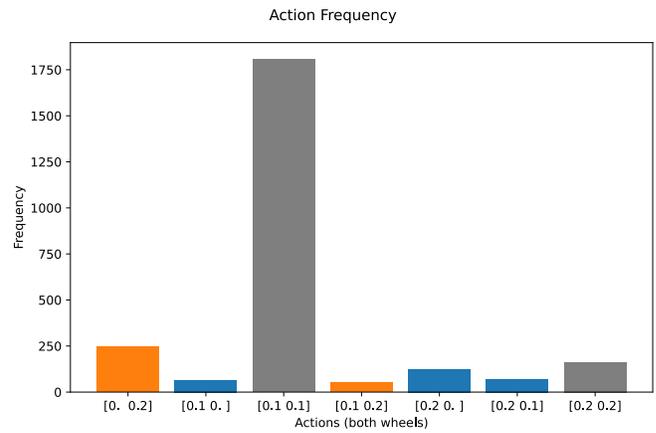}
  \caption{\label{fig:actions_1}
    Frequency of actions, which the agent takes if its policy is evaluated on the first racetrack (straight only), after learning the first sub-task.
  }
\end{figure}

\begin{figure}[!ht]
  \centering
  \includesvg[width=\linewidth]{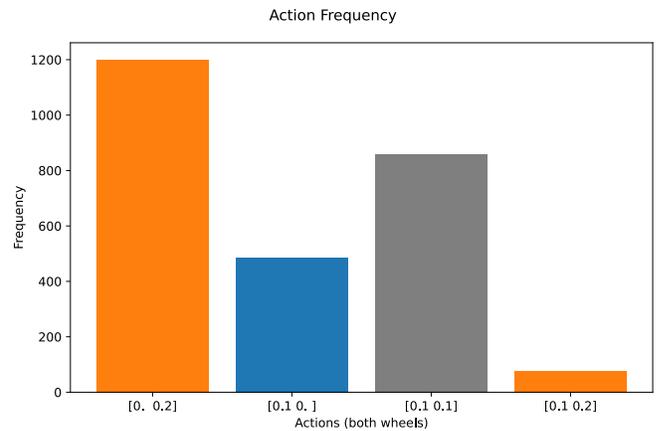}
  \caption{\label{fig:actions_2}
    Frequency of actions, which the agent takes if its policy is evaluated on the second racetrack (straight and left curve), after learning the first and second sub-task.
  }
\end{figure}

Frequencies of chosen actions are shown in \Cref{fig:actions_1,fig:actions_2}, where \cref{fig:actions_1} evaluates the first sub-task and \cref{fig:actions_2} the second one.
In both figures, all actions that have the same wheel speed (driving straight) are colored gray.
Actions which have a higher wheel speed on the right than on the left (perform a left curve) are shown in orange, and actions which have a higher wheel speed on the left than on the right (perform a right curve) are shown in blue.
\\
As expected, w.r.t. the first sub-task the agent takes straight actions the majority of time, so they are most likely for this track.
If the agent deviates from the line, other steering impulses are necessary, but this case is less common.
However, the number of diverse actions is still large, as $7/9$ are used.
During driving the second sub-task, frequencies are changing, because the racetrack has additionally sections of left curves.
Correspondingly, orange actions (performing a left curve) are more frequently chosen, than before.
To counteract still a possible overdrive, the frequency of blue actions (performing a right curve) is also increasing.
At the same time, it can be noted that the number of diverse actions decreases, by a kind of self-quantization of the action space.

\clearpage

{\small
  \bibliographystyle{IEEEtran}
  \bibliography{shorten.bib}
}

\end{document}